%% file: main.tex
\documentclass[10pt,twocolumn,letterpaper]{article}

\usepackage{booktabs} 
\usepackage{caption}  
\usepackage{amsmath}  
\usepackage{subfigure}
\usepackage{graphicx}
 \usepackage{multirow}
\usepackage[normalem]{ulem}
\useunder{\uline}{\ul}{}
\usepackage{iccv} 


\definecolor{iccvblue}{rgb}{0.21,0.49,0.74}
\usepackage[pagebackref,breaklinks,colorlinks,allcolors=iccvblue]{hyperref}

\def\paperID{1114} 
\def\confName{ICCV}
\def\confYear{2025}

\title{Intra-view and Inter-view Correlation Guided Multi-view Novel Class Discovery}

\author{
Xinhang Wan$^1$, Jiyuan Liu$^{2}$, Qian Qu$^1$, Suyuan Liu$^1$, Chuyu Zhang$^{3}$,\\ Fangdi Wang$^1$, Xinwang Liu$^{1,}$\thanks{Corresponding author}\space, En Zhu$^{1,*}$, Kunlun He$^4$
\\
$^1$College of Computer Science and Technology, National University of Defense Technology, Changsha, \\China
$^2$College of Systems Engineering, National University of Defense Technology, Changsha, China\\
$^3$ShanghaiTech University, Shanghai, China \quad 
$^4$Chinese PLA General Hospital, Beijing, China
\\
{\tt\small \{wanxinhang, xinwangliu, enzhu\}@nudt.edu.cn}\\
}

\begin{document}
\maketitle
\input{sec/0_abstract}    
\input{sec/1_intro}
\input{sec/2_related_work}
\input{sec/3_finalcopy}
\input{sec/4_optimization}
\input{sec/5_experiment}
\input{sec/6_conclusion}

\input{sec/7_acknowledgments}

{
    \small
    \bibliographystyle{ieeenat_fullname}
    \bibliography{main}
}


\end{document}

%% file: sec/0_abstract.tex
\begin{abstract}
In this paper, we address the problem of novel class discovery (NCD), which aims to cluster novel classes by leveraging knowledge from disjoint known classes. While recent advances have made significant progress in this area, existing NCD methods face two major limitations. First, they primarily focus on single-view data (e.g., images), overlooking the increasingly common multi-view data, such as multi-omics datasets used in disease diagnosis. Second, their reliance on pseudo-labels to supervise novel class clustering often results in unstable performance, as pseudo-label quality is highly sensitive to factors such as data noise and feature dimensionality. To address these challenges, we propose a novel framework named Intra-view and Inter-view Correlation Guided Multi-view Novel Class Discovery (IICMVNCD), which is the first attempt to explore NCD in multi-view setting so far.  Specifically, at the intra-view level, leveraging the distributional similarity between known and novel classes, we employ matrix factorization to decompose features into view-specific shared base matrices and factor matrices. The base matrices capture distributional consistency among the two datasets, while the factor matrices model pairwise relationships between samples. At the inter-view level, we utilize view relationships among known classes to guide the clustering of novel classes. This includes generating predicted labels through the weighted fusion of factor matrices and dynamically adjusting view weights of known classes based on the supervision loss, which are then transferred to novel class learning. Experimental results validate the effectiveness of our proposed approach.
\end{abstract}

%% file: sec/1_intro.tex
\section{Introduction}
\label{sec:intro}
Clustering, which aims to partition samples into groups based on their similarities \cite{chang2017deep, ren2024deep, dang2021nearest, liu2022deep,liu2023simple, murtagh2012algorithms,liu2022deep,yu2024zoo,10486880,10506102,liu2021multiview,wan2024contrastive,wan2024decouple,tu2025wage,gao2025generic,gao2024learning}, is a fundamental task in unsupervised learning and data mining. Traditional clustering methods, however, often assume no prior information and rely solely on sample similarities for partitioning \cite{nie2014clustering, jain2010data, hruschka2009survey,wen2025measure,Zhang_Lin_Yan_Yao_Wan_Li_Zhang_Ke_Xu_2025,zhang2025multi}. This assumption significantly diverges from human learning processes, where prior knowledge is commonly utilized to understand and classify new concepts. For example, a child familiar with phones and tablets can correctly distinguish new electronic devices, such as computers and smartwatches, by leveraging existing knowledge. Inspired by this cognitive ability, a novel learning paradigm, Novel Class Discovery (NCD), has been proposed to address this limitation \cite{zhong2021neighborhood, fini2021unified, joseph2022novel, zhao2022novel}.

NCD focuses on transferring knowledge from a labeled dataset to cluster an unlabeled dataset with non-overlapping categories \cite{roy2022class, hou2024nc2d}. This paradigm has shown great potential in applications such as disease discovery and face recognition. To the best of our knowledge, existing NCD methods can be broadly categorized into two types \cite{10376556}: two-stage and one-stage methods. Two-stage methods train a model on known classes and then fine-tune it for novel classes, while one-stage methods simultaneously learn representations for both classes and conduct the clustering process. One-stage methods generally achieve better performance because they can capture distributional relationships between the two datasets \cite{10203164}.

Despite significant advances, existing NCD methods face two major challenges that limit their practical application and performance: 1) Ignoring multi-view data: Most existing methods assume data comes from a single view, neglecting the increasingly common multi-view data. For example, in clinical diagnosis, accurately identifying new diseases (e.g., COVID-19) often requires multi-omics features, such as gene expression and imaging data, to work together for precise classification. Single-view data may not provide sufficient information for decision-making. Unlike single-view learning, multi-view settings \cite{10.5555/3298483.3298566,9305974,8387526, 7451227, jin2023deep,yu2025on,wang2023efficient,dong2025enhanced,dong2023cross,liu2024alleviate} require mining information from each view while effectively leveraging inter-view relationships to achieve consistent results. This limitation renders existing NCD methods unsuitable for handling multi-view data. 2) Dependency on pseudo-labels: Existing NCD methods often rely on pseudo-labels to supervise the learning of novel classes. However, the quality of pseudo-labels is easily affected by factors such as data distribution and feature space dimensionality, which can lead to unstable or even failed performance. Notably, these two issues are not independent. For instance, due to the diversity and complexity of multi-view data, generating pseudo-labels becomes more challenging and unreliable.

A key aspect of handling multi-view data is learning high-quality feature representations within each view while fusing information across views to achieve consistent results \cite{pmlr-v97-peng19a,9212617,wen2023unpaired,wang2024multiple,10243081,10325611,LiLiangTKDE23,LiLangTKDE24}. Inspired by this, we argue that NCD in multi-view settings can be addressed from two perspectives: From an intra-view perspective, it is essential to learn high-quality view-specific features by leveraging the data distributions of both known and novel sets. From an inter-view perspective, it is crucial to exploit correlations between views from known classes for feature fusion to obtain consistent representations for clustering. Fortunately, unlike traditional multi-view clustering that lacks prior information, the supervision signals from known classes can guide the learning of view weights for novel classes, making the weight assignment more reasonable and interpretable.

In light of this, we propose Intra-view and Inter-view Correlation Guided Multi-view Novel Class Discovery (IICMVNVD). Unlike existing one-stage NCD paradigms, our approach relies solely on the true labels of known classes for supervision, without involving the unreliable supervision of pseudo-labels. At the intra-view level, since the data distributions of both sets are similar, we learn a shared view-specific basis matrix via matrix factorization to capture the relationships between samples from known and novel classes, thereby obtaining better feature representations. At the inter-view level, we introduce two designs: 1) We optimize the view weights of novel classes based on the classification performance of known classes, enabling the model to dynamically adjust the importance of different views. 2) We directly guide the learning of view-consistent predicted labels using the relationships and data distributions of known and novel classes, rather than relying on pseudo-label supervision. The mutual guidance between the predicted labels of the two sets further boosts performance. Experimental results validate the effectiveness of our proposed method. Our contributions are summarized as follows:
\begin{itemize}
    \item To the best of our knowledge, this is the first attempt to address novel class discovery in the multi-view setting. Unlike existing pseudo-labeling methods, we predict the label by exploiting data distributions and class relationships across both novel and known categories.
    \item We exploit intra-view and inter-view correlations to transfer knowledge from known classes to novel classes. The intra-view shared basis matrix facilitates view-specific feature representation, while inter-view guidance helps capture the relationships between view weights and fusion strategies.
    \item Extensive experiments on diverse datasets demonstrate the effectiveness of our approach, achieving significant improvements over existing multi-view clustering and NCD methods.
\end{itemize}

%% file: sec/2_related_work.tex
\section{Related Work}
\label{sec:related_work}
In this section, we briefly introduce the most related research, including novel class discovery and multi-view clustering.
\subsection{Novel Class Discovery}
The primary goal of NCD is to transfer knowledge from known disjoint classes to cluster a novel unlabeled dataset. To the best of our knowledge, NCD was first studied in \cite{hsu2018learning,hsumulti}, and later formally defined by Han et al. in \cite{Han_2019_ICCV}, which spurred its rapid development. Unlike traditional clustering methods, NCD utilizes a labeled dataset as a reference for prior information when performing clustering. Furthermore, unlike semi-supervised learning, the labeled and unlabeled sets in NCD are disjoint in terms of classes.

As previously mentioned, NCD methods are mainly divided into two categories: two-stage and one-stage methods. Most early work focused on the two-stage framework. For example, \cite{hsu2018learning} first trains a prediction network on the labeled dataset and then applies it to the unlabeled dataset to calculate sample similarities, followed by clustering based on these similarities. In contrast, one-stage methods simultaneously learn from both labeled and unlabeled data through a joint optimization objective. For instance, the authors of \cite{hanautomatically} first pre-train feature representations via self-supervised learning on two datasets, then generate pseudo-labels using ranking statistics, and design a joint optimization objective to classify labeled data and cluster unlabeled data simultaneously. 
\begin{figure*}[htbp]
    \centering
    \includegraphics[width=0.98\textwidth]{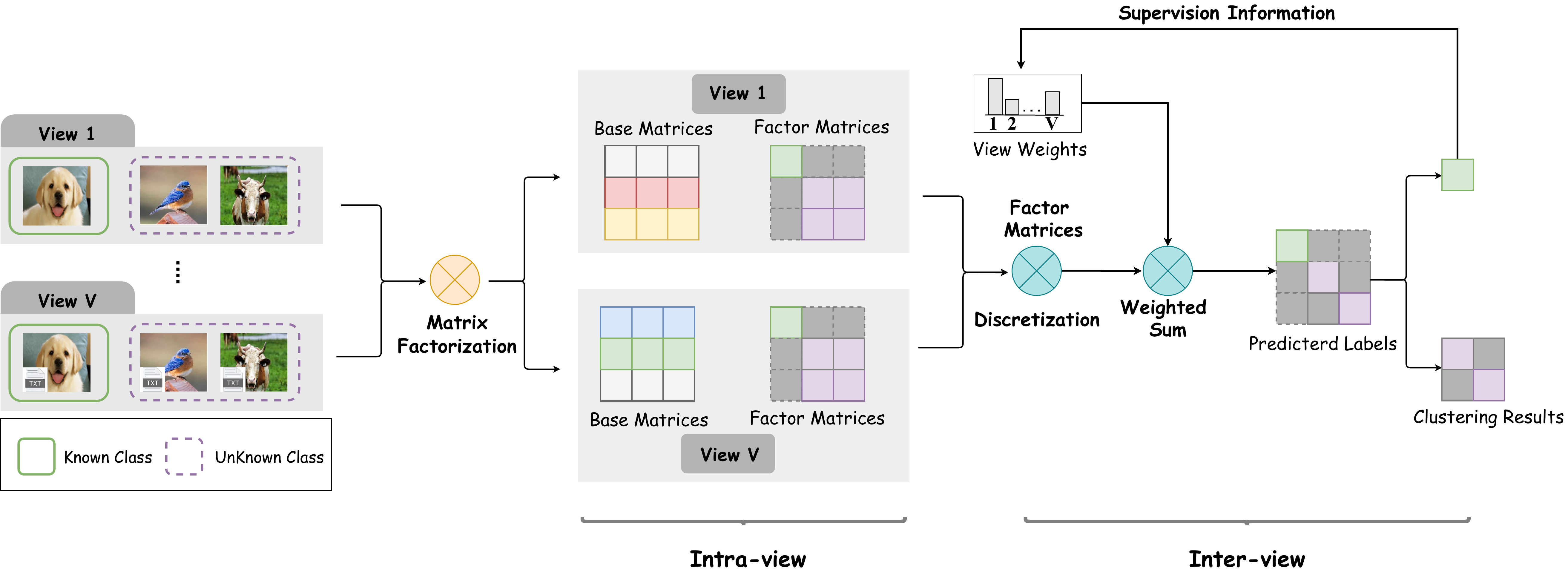}
    \caption{ Framework of the proposed algorithm: We first apply matrix factorization to decompose the features of the two sets into set-shared base matrices and factor matrices. The shared basis matrices effectively capture the intra-view distributional consistency between the two sets. Next, the factor matrices are further decomposed and discretized into predicted labels  with the incorporation of view weights. The discrepancy between the predicted and true labels of known classes is used to adjust the learning of view weights, enabling the guidance of inter-view relationships for unknown classes and facilitating cross-view fusion.}
    \label{alg_fig}
\end{figure*}
\subsection{Matrix Factorization-based Multi-view Clustering}
Multi-view clustering aims to partition samples by leveraging information from multiple views of data. With the increasing prevalence of multi-view data, this field has received significant attention in recent years \cite{2020Multi,10011211,10.1145/3474085.3475204,liu2023contrastive,yu2024towards,wen2025measure,Liu_2025_CVPR}. Among various approaches, matrix factorization-based multi-view clustering (MFMVC) \cite{liu2013multi,Gao2019MultiviewLM} has gained popularity due to its low computational complexity and ability to preserve essential multi-view information. This method typically decomposes features from all views into view-specific base matrices and a consensus factor matrix. The consensus matrix is then used for clustering to partition the multi-view data. For instance, the authors in \cite{8030316} proposed a diverse non-negative matrix factorization (NMF) method to enhance the diversity among multi-view representations. Liu et al. observed that separating clustering and multi-view feature learning could lead to suboptimal solutions. To address this, they further decomposed the consensus matrix into a consensus hard partition matrix and view-specific centroid matrices, directly learning the final clustering assignments.

However, existing algorithms often assume the absence of prior knowledge during clustering, which contrasts with human learning, where prior knowledge is crucial. Additionally, current NCD methods overlook the challenges of multi-view data, failing to effectively leverage information across views. Our work aims to develop a novel approach for multi-view novel class discovery (MVNCD) to address these limitations.

%% file: sec/3_finalcopy.tex
\section{Method}
\subsection{Overview}
In the MVNCD setting, we are given a labeled dataset 
$\mathcal{D}_l=\left\{\left\{\mathbf{x}^i_{v}\right\}_{v=1}^{V}, \mathbf{g}^i\right\}_{i=1}^{n_l}$ 
and an unlabeled dataset 
$\mathcal{D}_u=\left\{\left\{\mathbf{x}^j_{v}\right\}_{v=1}^{V}\right\}_{j=1}^{n_u}$, 
where $\mathbf{x}^i_{v}$ and $\mathbf{x}^j_{v}$ denote samples from the $v$-th view 
with feature dimension $d_v$. The categories in $\mathcal{D}_l$ and $\mathcal{D}_u$ are disjoint. The goal is to partition the samples in $\mathcal{D}_u$ into $k_u$ clusters 
by leveraging the knowledge extracted from $\mathcal{D}_l$ with $k_l$ categories. Consistent with most existing NCD settings, 
the number of clusters $k_u$ is assumed to be known a priori. At test time, we evaluate the clustering performance 
on the unlabeled dataset $\mathcal{D}_u$.

To address this issue, we utilize intra-view and inter-view correlations between known and novel classes to better partition the unlabeled dataset. From the intra-view perspective, we perform matrix factorization on the feature matrix of each view to learn a shared view-specific basis matrix for both datasets, which effectively enhances the quality of feature representations (see Sec. \ref{Intra_view_sec}). From the inter-view perspective, we leverage the relationships between views in the known classes to guide the learning of view relationships for the novel classes. Specifically, based on the factor matrices obtained in the intra-view step, we predict view-consistent labels by considering the data distribution and correlation across views. To improve clustering performance, we dynamically adjust view weights by combining the loss between the predicted and true labels of the known classes. The learned view weights are then applied to the novel classes, ensuring a more effective partitioning of the unlabeled dataset (see Sec. \ref{Inter_view_sec}). Finally, we provide the final loss function to jointly optimize intra-view and inter-view objectives (see Sec. \ref{loss_sec}). The basic framework of our method is given in Fig. \ref{alg_fig}.

\subsection{Intra-view Information Extraction}
\label{Intra_view_sec}
In this section, we discuss how to utilize the intra-view relationships to extract higher-quality features, which facilitates subsequent cross-view feature fusion. Traditional NCD algorithms often focus solely on the relationships within each dataset, neglecting the core assumption of NCD: the data distributions of the labeled and unlabeled datasets are similar. Based on this assumption, we propose to decompose the feature matrix of each view into a view-specific shared basis matrix $\mathbf{W}_v \in \mathbf{R}^{d_{v} \times k}$ and a factor matrix $\mathbf{Z}_v \in \mathbf{R}^{k \times n}$, where $k=k_l+k_u$ and $n=n_l+n_u$. 

The shared basis matrix $\mathbf{W}_v$ is designed to capture the distributional consistency across the labeled and unlabeled datasets, effectively modeling the common feature space shared by the two datasets. This ensures that the representation is not biased towards the labeled data while still being generalizable to the unlabeled data. On the other hand, the factor matrix $\mathbf{Z}_v$ models the relationships among individual samples. The formulation is provided as follows:

\begin{equation}
\begin{aligned}
\label{intra_eq}
&\min_{\mathbf{W}_v,\mathbf{Z}_v} \left\|\mathbf{X}_{v}-\mathbf{W}_v \mathbf{Z}_v \right\|_{F}^{2} \text { s.t. } 
\mathbf{W}_v^{\top} \mathbf{W}_v=\mathbf{I}_{k}.
\end{aligned}
\end{equation}
where $\mathbf{X}_{v}=[
\mathbf{X}_{v}^l,\mathbf{X}_{v}^u
] \in \mathbf{R}^{d_v \times n}$ is the feature matrix concatenated of the two sets in $v$-th view.

This formulation ensures that the basis matrix $\mathbf{W}_v$ remains orthonormal, which not only stabilizes the optimization process but also prevents redundancy.
\subsection{Inter-view Information Extraction}
\label{Inter_view_sec}
Unlike traditional NCD methods that consider single-source data only, MVNCD requires effectively leveraging multi-view information for fusion and decision-making. In practical applications, the quality and importance of different views often vary. For example, in medical diagnosis, multi-view data such as imaging and ultrasound information may both contribute to decision-making, but their importance differs depending on the disease being detected. Therefore, how to utilize the relationships between views for fusion is a critical aspect of MVNCD. Fortunately, with the presence of the labeled dataset, we can leverage supervised learning to capture view relationships in the known classes and transfer this knowledge to the learning of the unlabeled data.

Specifically, we first utilize the factor matrices obtained in Sec. \ref{Intra_view_sec} to derive the view-specific centroid matrix $\mathbf{A}_v$ and consistent predicted labels $\mathbf{Y}$ based on the k-means idea. Considering that the importance of each view varies, we assign a learnable weight to different views when generating consistent labels. The learning of view weights is constrained by the supervised loss on the known classes. In particular, we dynamically adjust the view weights $\boldsymbol{\alpha}$ by incorporating the difference between the predicted labels $\mathbf{Y}_l$ and the ground truth $\mathbf{G}_l$ of the known classes, and then apply this optimization strategy to the view weight learning for the novel classes. The overall optimization objective is as follows:
\begin{equation}
\begin{aligned}\label{inter_loss_eq}
&\min_{\boldsymbol{\alpha},\mathbf{W}_v,\mathbf{A}_v, \mathbf{Y}} \sum_{v=1}^{V}{\alpha_v}^2\left\|\mathbf{X}_{v}-\mathbf{W}_v \mathbf{A}_v \mathbf{Y}\right\|_{F}^{2}+ \lambda_1\left\|\mathbf{Y}_l-\mathbf{G}_l\right\|_{F}^{2}\\
&\text { s.t. } 
\boldsymbol{\alpha}^{\top} \mathbf{1}=1, \boldsymbol{\alpha}\geq\mathbf{0}, \mathbf{W}_v^{\top} \mathbf{W}_v=\mathbf{I}_{k}, \mathbf{Y} \in \Phi^{k \times n}.
\end{aligned}
\end{equation}
where $\mathbf{Y}=[
\mathbf{Y}_l,\mathbf{Y}_u
] \in \mathbf{R}^{k \times n}$ denotes the matrix of predicted labels, where each column is a $k$-dimensional one-hot vector representing the predicted label of a corresponding sample.

Through this design, the supervised loss on the known classes not only regulates the generation of $\mathbf{Y}_l$ but also contributes to the learning of view weights.

\subsection{Loss Function}
\label{loss_sec}
Due to the similarity in data distributions between the known and novel classes, there is a risk that novel class samples may be misclassified into the known classes during the joint learning process. To address this issue, we impose constraints on the labels of the novel classes to ensure label disjointness between the two sets. Building upon Eq. \eqref{inter_loss_eq}, we propose the final objective function as follows:

\begin{equation}
\begin{aligned}\label{final_loss}
&\min_{\boldsymbol{\alpha},\mathbf{W}_v,\mathbf{A}_v, \mathbf{Y}} \sum_{v=1}^{V}{\alpha_v}^2\left\|\mathbf{X}_{v}-\mathbf{W}_v \mathbf{A}_v \mathbf{Y}\right\|_{F}^{2} + \\
&\lambda_1\left\|\mathbf{Y}_l-\mathbf{G}_l\right\|_{F}^{2} - \lambda_2\sum_{\mathbf{g}^{i} \in \mathbf{G}^l} \sum_{\mathbf{y}^{j} \in \mathbf{Y}^u} \left\|\mathbf{g}^{i}-\mathbf{y}^{j}\right\|_{F}^{2}\\
&\text { s.t. } 
\boldsymbol{\alpha}^{\top} \mathbf{1}=1, \boldsymbol{\alpha}\geq\mathbf{0}, \mathbf{W}_v^{\top} \mathbf{W}_v=\mathbf{I}_{k}, \mathbf{Y} \in \Phi^{k \times n},
\end{aligned}
\end{equation}
where $\lambda_1$ and $\lambda_2$ are hyperparameters that control the trade-offs among the terms. 

Through this design, the loss function not only optimizes intra-view and inter-view feature representations but also ensures the separation of labels between known and novel classes, thereby improving the clustering performance on unlabeled data.

%% file: sec/4_optimization.tex
\section{Optimization}
We develop a four-step alternating optimization algorithm to solve the resultant problem in Eq.~\eqref{final_loss}. In each step, we optimize one variable while keeping the others fixed.

\subsection{\texorpdfstring{$\mathbf{W}_v$}{Wv} Subproblem}
With the other variables fixed in Eq.~\eqref{final_loss}, $\mathbf{W}_v$ can be updated by solving the following formula:
\begin{equation}
\begin{aligned}\label{opt_w}
\max_{\mathbf{W}_v} \operatorname{Tr}\left(\mathbf{W}_v^{\top} \mathbf{B}_v\right), \quad \text{s.t. } 
\mathbf{W}_v^{\top} \mathbf{W}_v = \mathbf{I}_{k},
\end{aligned}
\end{equation}
where $\mathbf{B}_v = \mathbf{X}_v \mathbf{Y}^{\top} \mathbf{A}_v^{\top}$.

Suppose the matrix $\mathbf{B}_v$ has the singular value decomposition (SVD) form as $\mathbf{B}_v = \mathbf{S}_v \boldsymbol{\Sigma}_v \mathbf{V}_v^{\top}$. The optimization problem in Eq.~\eqref{opt_w} can then be solved by the closed-form solution given in \cite{ijcai2019-524} as:
\begin{equation}\label{get_hpv}
\begin{aligned}
\mathbf{W}_v = \mathbf{S}_v \mathbf{V}_v^{\top}.
\end{aligned}
\end{equation}

\subsection{\texorpdfstring{$\mathbf{A}_v$}{Av} Subproblem}
Fixing other variables, the formula in Eq.~\eqref{final_loss} can be reduced to:
\begin{equation}
\begin{aligned}\label{opt_A}
\min_{\mathbf{A}_v} \left\|\mathbf{X}_v - \mathbf{W}_v \mathbf{A}_v \mathbf{Y}\right\|_{F}^{2}.
\end{aligned}
\end{equation}

By setting the derivative to zero, the solution is:
\begin{equation}
\begin{aligned}
\mathbf{A}_v = \mathbf{W}_v^{\top} \mathbf{X}_v \mathbf{Y}^{\top} \left(\mathbf{Y} \mathbf{Y}^{\top}\right)^{-1}.
\end{aligned}
\end{equation}

\subsection{\texorpdfstring{$\mathbf{Y}$}{Y} Subproblem}
Since $\mathbf{Y}_l$ and $\mathbf{Y}_u$ are independent of each other for optimization, we further divide the problem into two subproblems.

\subsubsection{\texorpdfstring{$\mathbf{Y}_l$}{Yl} Subproblem}
Fixing the other variables, the optimization problem for $\mathbf{Y}_l$ can be rewritten as:
\begin{equation}
\begin{aligned}\label{opt_Yl}
&\min_{\mathbf{Y}_l} \sum_{v=1}^{V} {\alpha_v}^2 \left\|\mathbf{X}_v^l - \mathbf{M}_v \mathbf{Y}_l\right\|_{F}^{2} + \lambda_1 \left\|\mathbf{Y}_l - \mathbf{G}_l\right\|_{F}^{2}, \\
&\text{s.t. } \mathbf{Y}_l \in \Phi^{k \times n_l},
\end{aligned}
\end{equation}
where $\mathbf{M}_v = \mathbf{W}_v \mathbf{A}_v$.

The optimal solution of $\mathbf{y}^i \in \mathbf{Y}_l$ can be obtained by solving:
\begin{equation}
\begin{aligned}
&\min_{\mathbf{y}^i} {\mathbf{y}^i}^{\top}\left(\sum_{v=1}^{V} {\alpha_v}^2 \mathbf{M}_v^{\top} \mathbf{M}_v\right)\mathbf{y}^i - 2\sum_{v=1}^{V} {\alpha_v}^2 {\mathbf{x}_v^i}^{\top} \mathbf{M}_v \mathbf{y}^i \\
&\quad - 2\lambda_1 {\mathbf{g}^i}^{\top} \mathbf{y}^i, \quad \text{s.t. } \mathbf{y}^i \in \{0, 1\}^{k \times 1}, \mathbf{y}^i \mathbf{1} = 1.
\end{aligned}
\end{equation}

Let $\mathbf{b} = \operatorname{diag}\left(\sum_{v=1}^{V} {\alpha_v}^2 \mathbf{M}_v^{\top} \mathbf{M}_v\right)$ and $\mathbf{c} = \sum_{v=1}^{V} {\alpha_v}^2 \mathbf{M}_v^{\top} \mathbf{x}_v^i + \lambda_1 \mathbf{g}^i$. The position of the non-zero element is:
\begin{equation}
\arg \min_l \mathbf{b}_l - 2\mathbf{c}_l.
\end{equation}

\subsubsection{\texorpdfstring{$\mathbf{Y}_u$}{Yu} Subproblem}
Similarly, the optimization problem for $\mathbf{Y}_u$ can be rewritten as:
\begin{equation}
\begin{aligned}\label{opt_Yu}
&\min_{\mathbf{Y}_u} \sum_{v=1}^{V} {\alpha_v}^2 \left\|\mathbf{X}_v^u - \mathbf{M}_v \mathbf{Y}_u\right\|_{F}^{2} - \lambda_2 \sum_{\mathbf{g}^i \in \mathbf{G}_l} \sum_{\mathbf{y}^j \in \mathbf{Y}_u} \left\|\mathbf{g}^i - \mathbf{y}^j\right\|_{F}^{2}, \\
&\text{s.t. } \mathbf{Y}_u \in \Phi^{k \times n_u}.
\end{aligned}
\end{equation}

The optimal solution of $\mathbf{y}^j \in \mathbf{Y}_u$ can be obtained by solving:
\begin{equation}
\begin{aligned}
&\min_{\mathbf{y}^j} {\mathbf{y}^j}^{\top}\left(\sum_{v=1}^{V} {\alpha_v}^2 \mathbf{M}_v^{\top} \mathbf{M}_v\right)\mathbf{y}^j - 2\sum_{v=1}^{V} {\alpha_v}^2 {\mathbf{x}_v^j}^{\top} \mathbf{M}_v \mathbf{y}^j \\
&\quad + 2\lambda_2 \sum_{\mathbf{g}^i \in \mathbf{G}_l} {\mathbf{g}^i}^{\top} \mathbf{y}^j, \quad \text{s.t. } \mathbf{y}^j \in \{0, 1\}^{k \times 1}, \mathbf{y}^j \mathbf{1} = 1,
\end{aligned}
\end{equation}
which is similar to the optimization problem of $\mathbf{Y}_l$.
\subsection{\texorpdfstring{$\boldsymbol{\alpha}$}{Alpha} Subproblem}
Finally, given other variables, the formulation about $\boldsymbol{\alpha}$ can be solved via optimizing:
\begin{equation}\label{opt_alpha}
\begin{aligned}
\min \sum_{v=1}^{V} \alpha_v^2 r_v^2, \quad \text{s.t. } \boldsymbol{\alpha}^{\top} \mathbf{1} = 1, \boldsymbol{\alpha} \geq \mathbf{0},
\end{aligned}
\end{equation}
where
\begin{equation}
\begin{aligned}
r_v^2 = \left\|\mathbf{X}_v - \mathbf{W}_v \mathbf{A}_v \mathbf{Y}\right\|_{F}^{2}.
\end{aligned}
\end{equation}

Based on the Cauchy-Schwarz inequality, $\alpha_v$ is updated as:
\begin{equation}\label{get_alpha}
\begin{aligned}
\alpha_v = \frac{\frac{1}{r_v^2}}{\sum_{v=1}^{V} \frac{1}{r_v^2}}.
\end{aligned}
\end{equation}

The overview of the alternate optimization strategy is outlined in Algorithm 1.

\begin{algorithm}[h]
\renewcommand{\algorithmicrequire}{\textbf{Input:}}
\renewcommand{\algorithmicensure}{\textbf{Output:}}
    \caption{IICMVNCD}
    \label{algo}
    \begin{algorithmic}[1]
        \REQUIRE The labeled dataset $\mathcal{D}_l$, unlabeled dataset $\mathcal{D}_u$ with $k_u$ clusters, hyper-parameters $\lambda_1$ and $\lambda_2$.
        \ENSURE Cluster partition matrix $\mathbf{Y}_u$.
        \STATE Initialize  $\boldsymbol{\alpha}$, $\mathbf{W}_v$, $\mathbf{A}_v$ and $\mathbf{Y}$.
   \WHILE{not converged}
    
         \STATE  Update $\mathbf{W}_v$ by solving Eq. \eqref{opt_w};
         \STATE  Update $\mathbf{A}_v$ by solving Eq. \eqref{opt_A};
         \STATE  Update $\mathbf{Y}$ by solving Eq. \eqref{opt_Yl} and \eqref{opt_Yu};
         \STATE  Update $\mathbf{\alpha}$ by solving Eq. \eqref{opt_alpha};
     \ENDWHILE
     
    \end{algorithmic}
\end{algorithm}
\subsection{Theoretical Analysis}
\subsubsection{Convergence}
Most existing NCD methods fail to provide theoretical guarantees for convergence. In contrast, our method is theoretically proven to converge. For simplicity of expression, we reformulate the objective function in Eq.~\eqref{final_loss} as:
\begin{equation}
\begin{aligned}
\min_{\mathbf{W}_v, \mathbf{A}_v, \mathbf{Y}, \boldsymbol{\alpha}} \mathcal{J}\left(\mathbf{W}_v, \mathbf{A}_v, \mathbf{Y}, \boldsymbol{\alpha}\right).
\end{aligned}
\end{equation}

Our optimization process consists of four steps in each iteration. Let the superscript $t$ represent the optimization variables at the $t$-th iteration. We have:
\begin{equation}\label{convergence_eq}
\begin{aligned}
&\mathcal{J}\left(\mathbf{W}_v^{(t+1)}, \mathbf{A}_v^{(t+1)}, \mathbf{Y}^{(t+1)}, \boldsymbol{\alpha}^{(t+1)}\right) \leq \\
&\mathcal{J}\left(\mathbf{W}_v^{(t)}, \mathbf{A}_v^{(t)}, \mathbf{Y}^{(t)}, \boldsymbol{\alpha}^{(t)}\right),
\end{aligned}
\end{equation}
which demonstrates that the loss monotonically decreases during the optimization process.

Furthermore, for the loss function in Eq.~\eqref{final_loss}, we can derive the following inequality:
\begin{equation}
\begin{aligned}
\sum_{\mathbf{g}^{i} \in \mathbf{G}^l} \sum_{\mathbf{y}^{j} \in \mathbf{Y}^u} \left\|\mathbf{g}^{i} - \mathbf{y}^{j}\right\|_{F}^{2} \leq n_l n_u \sqrt{2}.
\end{aligned}
\end{equation}

Hence, Eq.~\eqref{final_loss} has a lower bound satisfying:
\begin{equation}
\begin{aligned}
\mathcal{J}\left(\mathbf{W}_v, \mathbf{A}_v, \mathbf{Y}, \boldsymbol{\alpha}\right) \geq - \lambda_2 n_l n_u \sqrt{2}.
\end{aligned}
\end{equation}

Therefore, the proposed method is theoretically guaranteed to converge.

\subsubsection{Time Complexity}
The optimization process comprises four subproblems, and we analyze the time complexity of each subproblem separately per iteration. In the $\mathbf{W}_v$ subproblem, it costs $\mathcal{O}\left(d_vnk\right)$ to compute $\mathbf{B}_v$. Then, it requires $\mathcal{O}\left(d_vk^2\right)$ to perform SVD on it. Thus, the total cost of updating $\left\{\mathbf{W}_v\right\}_{v=1}^V$ is $\mathcal{O}\left(d(nk+k^2)\right)$, where $d=\sum_{v=1}^V d_v$. In the $\mathbf{A}_v$ subproblem, it takes $\mathcal{O}\left(d_vnk+d_vk^2+k^3\right)$ to solve for $\mathbf{A}_v$ in each view. Therefore, the total complexity of updating $\left\{\mathbf{A}_v\right\}_{v=1}^V$ is $\mathcal{O}\left(dnk+dk^2+Vk^3\right)$. For the $\mathbf{Y}$ subproblem, the cost of updating $\mathbf{Y}$ is $\mathcal{O}\left(dnk\right)$. Finally, for the $\boldsymbol{\alpha}$ subproblem, the update requires $\mathcal{O}\left(k\right)$. 

Combining all subproblems, the overall time complexity per iteration is:
\begin{equation}
\mathcal{O}\left(d(nk+k^2)+Vk^3\right),
\end{equation}
which is linear with respect to the sample number $n$. This ensures the scalability of the proposed method.

%% file: sec/5_experiment.tex
\section{Experiment}
\subsection{Experimental Setting}
\subsubsection{Datasets}
We utilized eight multi-view datasets in our experiments, including BRCA, KIPAN, uci-digit, Cora, Wiki, CCV, STL10, and YTB10. Their information is summarized in Table \ref{tab_dataset}. Among these, BRCA and KIPAN are multi-omics datasets. More detailed information is provided in the Appendix due to space limitations.

For the aforementioned datasets, we divide the known classes and novel classes in a 1:1 ratio. For simplicity, we assume that the first half of the classes are designated as known classes, while the second half are treated as novel classes. If the total number of classes is odd, the number of known classes is set to be one less than that of the unknown classes by default.

\begin{table}[h]
    \begin{center}
        \caption{Datasets used in our experiments.} 
        \label{tab_dataset}
        \begin{tabular}{ c c c c}
            \toprule
            Dataset & Samples & Views & Clusters \\  \hline
            BRCA & 511 & 3 & 4 \\  
            KIPAN & 707 & 3 & 3 \\  
            uci-digit & 2,000 & 3 & 10 \\  
            Cora & 2,708 & 4 & 7 \\  
            Wiki & 2,866 & 2 & 10 \\  
            CCV & 6,773 & 3 & 20 \\  
            STL10 & 13,000 & 4 & 10 \\  
            YTB10 & 38,654 & 4 & 10 \\  \bottomrule
        \end{tabular}
    \end{center}
\end{table}
\begin{table*}[]
\caption{Comparison of ACC, NMI, and Purity for various MVC and NCD algorithms across eight benchmark datasets. The best results are highlighted in bold, while the second-best results are underlined.}\label{our_main_table}
\centering
\begin{tabular}{|c|cccccc|ccc|}
\hline
                           & \multicolumn{6}{c|}{MVC Methods}                                                                                                                 & \multicolumn{3}{c|}{NCD Methods} \\
\cline{2-10} 
\multirow{-2}{*}{Datasets} & LMVSC                        & FPMVS                        & OPMC                         & {\color[HTML]{333333} AWMVC} & RCAGL  & AEVC        & IIC         & CKD               & Ours           \\
\hline
\multicolumn{10}{|c|}{ACC}                                                                                                                                                                                                        \\
\hline
BRCA                       & {\ul 98.18} & {\color[HTML]{333333} 80.61} & {\color[HTML]{333333} 52.73} & {\color[HTML]{333333} 51.02} & 56.36 & 86.67       & 56.22       & 84.32       & \textbf{98.79} \\
KIPAN                      & 87.68                        & 91.58                        & 89.86                        & 80.46                        & 51.95 & 91.67       & 83.78       & {\ul 91.73}       & \textbf{92.51} \\
uci-digit                  & 89.82                        & 87.40                        & {\ul 93.60}                        & 57.09                        & 87.50 & 92.60       & 77.05       & 92.50       & \textbf{95.30} \\
Cora                       & 57.36                        & 55.93                        & 42.33                        & 40.58                        & 0.00  & {\ul 63.20} & 48.56       & 35.23             & \textbf{76.36} \\
Wiki                       & {\ul 64.31}                  & 39.70                        & 30.77                        & 21.97                        & 33.93 & 37.30       & 32.77       & 61.05             & \textbf{65.42} \\
CCV                        & 32.17                        & 33.30                        & 31.20                        & 22.55                        & 27.06 & {\ul 33.52} & 26.09       & 31.66             & \textbf{34.20} \\
STL10                      & 61.91                        & 85.90                        & {\ul 98.95}                  & 60.34                        & 80.12 & 98.74       & 71.16       & 96.11             & \textbf{99.02} \\
YTB10                      & 84.31                        & 92.05                        & 79.08                        & 57.98                        & 75.01 & 91.59       & 72.84       & {\ul 93.01}       & \textbf{94.55} \\
\hline
\multicolumn{10}{|c|}{NMI}                                                                                                                                                                                                        \\
\hline
BRCA                       & {\ul 87.81}                  & 37.39                        & 0.01                         & 37.54                        & 6.46  & 50.05       & 35.47       & 86.93             & \textbf{90.45} \\
KIPAN                      & 46.27                        & 58.42                        & 55.00                        & 52.60                        & 0.05  & {\ul 61.08} & 55.99       & 58.80             & \textbf{62.02} \\
uci-digit                  & 83.85                        & 79.89                        & 84.99                        & 69.14                        & 74.43 & 82.21       & 73.45       & {\ul 85.61}       & \textbf{86.59} \\
Cora                       & 25.19                        & 21.97                        & 10.95                        & 16.16                        & 0.00  & {\ul 29.96} & 24.01       & 2.12              & \textbf{44.59} \\
Wiki                       & {\ul 49.98}                  & 14.01                        & 6.06                         & 8.48                         & 9.99  & 13.95       & 14.04       & 35.10             & \textbf{66.44} \\
CCV                        & 17.26                        & 16.62                        & {\ul 17.81}                  & 16.66                        & 12.62 & 16.83       & 16.70       & 16.83             & \textbf{19.24} \\
STL10                      & 56.90                        & 72.39                        & {\ul 96.01}                  & 62.45                        & 63.50 & 95.32       & 62.37       & 87.28             & \textbf{96.22} \\
YTB10                      & 75.83                        & {\ul 83.08}                  & 72.89                        & 71.26                        & 70.66 & 81.14       & 77.55       & 82.84             & \textbf{86.85} \\
\hline
\multicolumn{10}{|c|}{Purity}                                                                                                                                                                                                        \\
\hline
BRCA                       & {\ul 98.18}                        & 80.61                        & 67.88                        & 91.64                        & 87.27 & 86.67       & 89.78       & 84.32       & \textbf{98.79} \\
KIPAN                      & 87.68                        & 91.58                        & 89.86                        & 88.89                        & 88.46 & 91.67       & 90.80       & {\ul 91.73}       & \textbf{92.51} \\
uci-digit                  & 90.65                        & 87.43                        & {\ul 93.60}                        & 91.34                        & 87.50 & 92.60       & 90.80       &  92.50      & \textbf{95.30} \\
Cora                       & 57.36                        & 55.93                        & 42.40                        & 49.38                        & 0.00  & {\ul 63.20} & 60.62       & 35.50             & \textbf{76.36} \\
Wiki                       & {\ul 65.11}                  & 45.32                        & 37.37                        & 41.17                        & 52.94 & 43.48       & 46.53       & 61.12             & \textbf{75.47} \\
CCV                        & 36.80                        & 35.73                        & 36.35                        & {\ul 39.12}                  & 38.77 & 37.31       & 37.22       & 36.44             & \textbf{39.95} \\
STL10                      & 64.09                        & 85.91                        & {\ul 98.95}                  & 82.79                        & 80.12 & 98.74       & 83.91       & 96.11             & \textbf{99.02} \\
YTB10                      & 86.08                        & 92.70                        & 79.08                        & 90.20                        & 82.99 & 91.60       & 92.23       & {\ul 93.01}       & \textbf{94.55} \\
\hline
\end{tabular}
\end{table*}
\subsubsection{Compared Algorithms and Experimental Setup}
We compare our proposed method with the six multi-view clustering  and two novel class discovery methods, and they are summarized as follows:

\begin{enumerate} 
    \item \textbf{LMVSC}\cite{LMVSC}: Performs spectral clustering on the final anchor graph derived for each view, achieving linear computational efficiency for large-scale datasets.

    \item \textbf{FPMVS}\cite{FPMVS-CAG}: Introduces a subspace clustering approach that simultaneously performs anchor selection and subspace graph construction, eliminating the need for additional hyper-parameters.

    \item \textbf{OPMC}\cite{OPMC}: A one-pass approach that seamlessly integrates multi-view matrix factorization and partition generation into a unified framework.

    \item \textbf{AWMVC}\cite{AWMVC}: Learns coefficient matrices from view-specific base matrices. It captures more comprehensive information by projecting original features into distinct low-dimensional spaces.

    \item \textbf{RCAGL}\cite{RCAGL}: Proposes a robust and consistent anchor graph learning method for multi-view clustering. It constructs a consistent anchor graph by simultaneously learning a consistent part to capture inter-view commonality and a view-specific part to filter out noise.

     \item \textbf{AEVC}\cite{AEVC}: A plug-and-play anchor enhancement strategy for multi-view clustering. By analyzing relationships between anchors and samples across neighboring views, it ensures consistency and alignment across similar views.

    \item \textbf{IIC}\cite{10203164}: A novel class discovery (NCD) method that models inter-class and intra-class constraints using symmetric Kullback-Leibler divergence.

    \item \textbf{CKD}\cite{10376556}: An NCD method that addresses the information loss of class relations in existing methods by utilizing a knowledge distillation framework.
\end{enumerate}

For our method, $\lambda_1$ and $\lambda_2$ are tuned from  $10 .^{\wedge}[0,1, \cdots, 5]$. As for the compared methods, the parameter settings of all baselines were tuned according to the recommendations provided in their respective papers. For NCD methods, originally designed for image datasets, we adapt them for multi-view datasets by replacing their backbone with $V$ separate $3$ -layer MLP networks. Each network acts as a feature extractor for one view, and the features from all views are concatenated to form the input for the subsequent NCD algorithms. The clustering performance is evaluated using standard metrics, including clustering accuracy (ACC), normalized mutual information (NMI), and Purity. All experiments were conducted on a machine equipped with an Intel Core i9-10850K CPU @ 3.60GHz, 64GB RAM, and an Nvidia RTX 3090 GPU. 

\subsection{Experimental Results}
To demonstrate the effectiveness of our method, we conducted experiments reporting the results of IICMVNCD on eight widely used benchmark datasets, comparing our algorithm with others. The results are presented in Table \ref{our_main_table}.

From the table, we observe that our algorithm consistently outperforms both MVC and NCD baselines on every dataset, demonstrating its effectiveness. Relative to MVC methods, our approach exploits prior knowledge from known classes, enabling a better grasp of the data distribution and inter‑view relationships for the unseen classes and thus yielding superior clustering results. Existing NCD methods, on the other hand, are largely designed for single‑view data and do not effectively exploit complementary cues across views; as a consequence, their performance is generally inferior—often even worse than MVC baselines. These findings highlight the importance of developing NCD techniques that can fully leverage multi‑view information, as accomplished by our proposed framework. These findings highlight the importance of developing NCD techniques that can fully leverage multi‑view information, as accomplished by our proposed framework.

\subsection{Convergence}
\begin{figure}[htbp]
	\centering
	\vspace{-0.2cm}
        \subfigure{\includegraphics[width=0.238\textwidth]{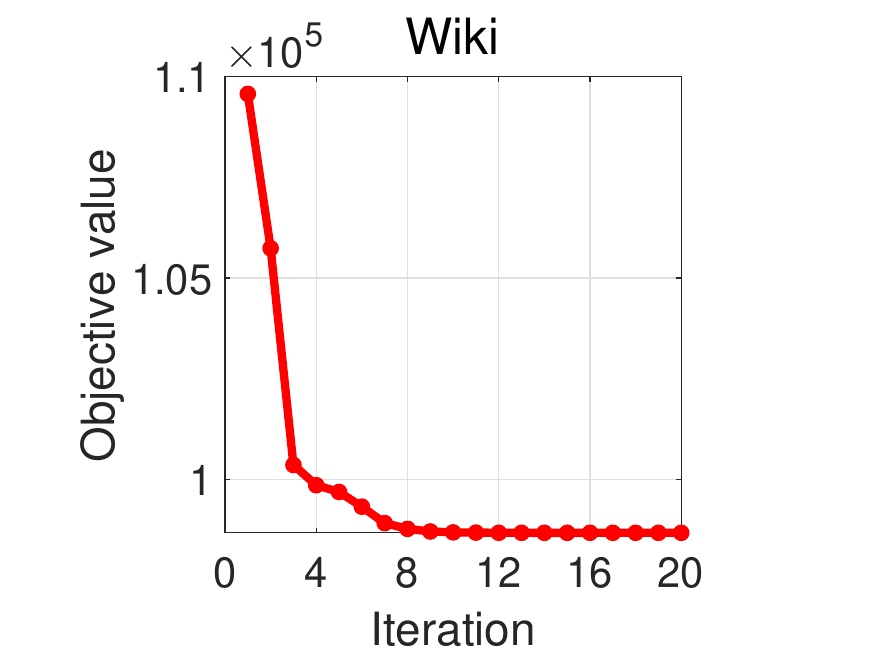}}
	\hspace{-0.2cm}
	\subfigure{
		\includegraphics[width=0.238\textwidth]{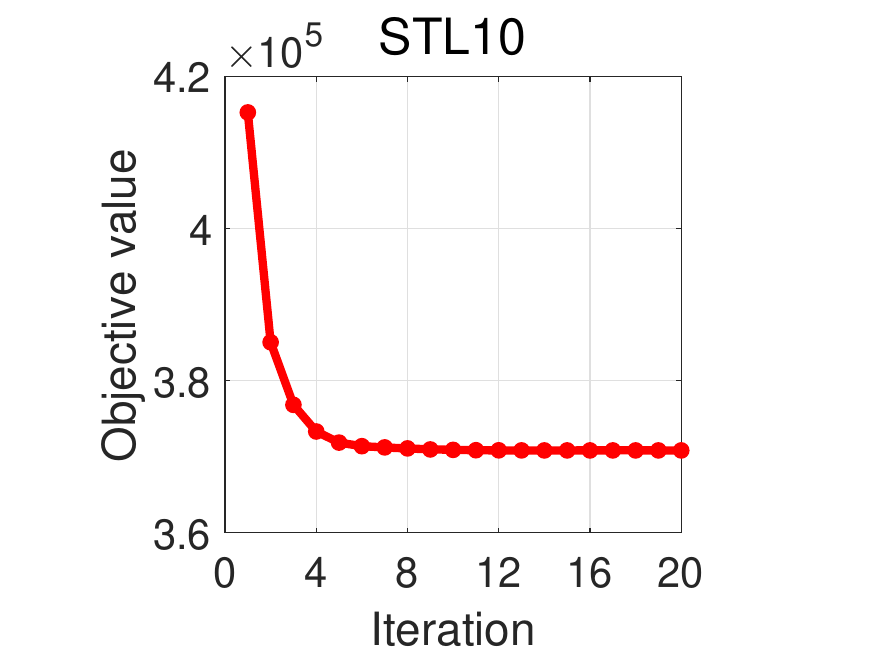}}
	\hspace{-0.2cm}
	\caption{The objective values of our method vary with iterations on two datasets.}
	\label{fig_iter}
\end{figure}

As proved above, our proposed algorithm is theoretically convergent. To confirm its practical convergence, we plot the changes in the objective value of our method across iterations on two datasets, as illustrated in Figure \ref{fig_iter}. From the figure, it is evident that the objective value decreases monotonically and converges within fewer than fifty iterations.
\begin{table}[]
\caption{The ablation study to investigate the effectiveness of main modules of our method in terms of ACC. The best results are marked in bold.}
\label{ablation_exp}
\centering
\begin{tabular}{ccccc}
\toprule
Datasets    & w/o $\boldsymbol{\alpha}$ and $\mathcal{D}_l$ & w/o $\boldsymbol{\alpha}$ & w/o $\mathcal{D}_l$ & Ours           \\
\hline
BRCA      & 52.73                                         & 58.79                     & 58.79               & \textbf{98.79} \\
KIPAN     & 89.86                                         & 90.06                     & 90.95               & \textbf{92.51} \\
uci-digit & 93.60                                         & 95.20                     & 93.80               & \textbf{95.30} \\
Cora      & 42.33                                         & 56.66                     & 42.20               & \textbf{76.36} \\
Wiki      & 30.77                                         & 35.44                     & 64.35               & \textbf{65.42} \\
CCV       & 31.20                                         & 33.41                     & 32.78               & \textbf{34.20} \\
STL10     & 98.95                                         & 99.02                     & 98.95               & \textbf{99.02} \\
YTB10     & 79.08                                         & 94.32                     & 79.08               & \textbf{94.55} \\ \bottomrule
\end{tabular}
\end{table}

\subsection{Ablation Study}
In this section, we conduct ablation experiments to validate the effectiveness of the proposed modules. In our framework, we primarily leverage information from known classes to enhance the clustering of novel classes in two key ways. First, we utilize the similarity between the distributions of known and unknown classes at the intra-view level through matrix factorization, allowing us to learn shared view-specific basis matrices and factor matrices that capture the relationships between samples of known and unknown classes. Second, at the inter-view level, we use the view weights learned from known classes to guide the weight learning for unknown classes. To evaluate the contributions of these components, we perform experiments by utilizing only the distribution of novel classes during matrix factorization while retaining the view weight guidance module (w/o $\mathcal{D}_l$). Additionally, we assess the scenario where we use only the information from unknown classes for guiding view weight learning while keeping the intra-view component (w/o $\boldsymbol{\alpha}$). Finally, we also consider the case where both modules are removed (w/o $\boldsymbol{\alpha}$ and $\mathcal{D}_l$) and report the results in Table \ref{ablation_exp}. From the table, it is evident that removing any one of the modules results in a decline in the performance of our model, thereby showing the usefulness of each module.

\subsection{Parameter Sensitivity}
\begin{figure}[htbp]
	\centering
	\vspace{-0.2cm}
        \subfigure{	\includegraphics[width=0.238\textwidth]{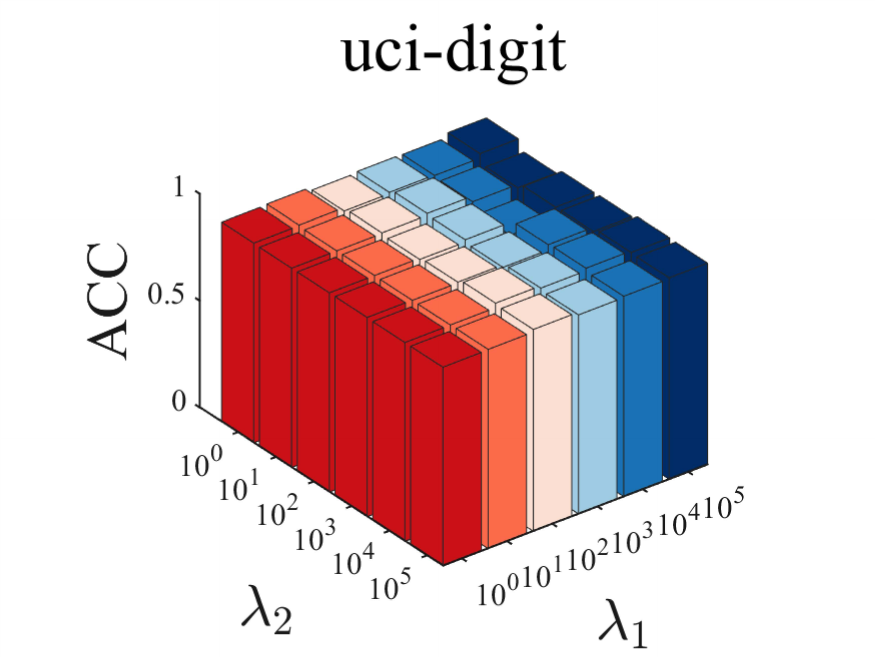}}
	\hspace{-0.2cm}
        \subfigure{\includegraphics[width=0.238\textwidth]{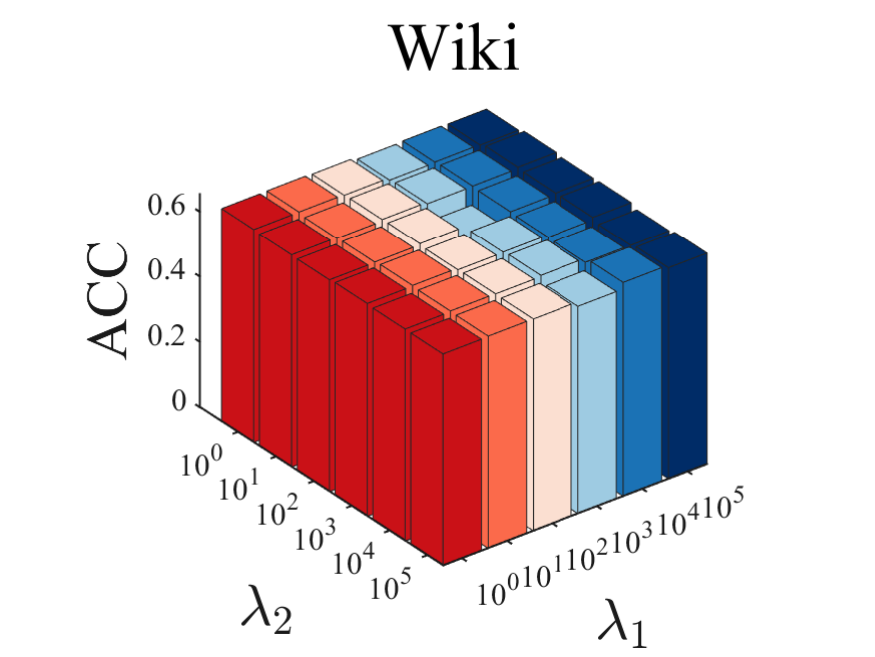}}
	\hspace{-0.2cm}
	\caption{The parameter sensitivity of our method on uci-digit and Wiki.}
	\label{fig_lambda}
\end{figure}

We employed two hyperparameters to adjust the importance of each term in Eq. \eqref{final_loss}. To investigate the selection of hyperparameters and their impact on the final results, we plotted the effects of hyperparameter fluctuations on ACC in Fig. \ref{fig_lambda}. It can be observed that our results show minimal variation with changes in hyperparameters, which shows the stability of our method. Moreover, in practical applications, due to the presence of known class label information, we can also adjust the hyperparameters for unknown classes based on the model's performance on known classes. We will explore how to tune them based on the results of known classes in future research.

%% file: sec/6_conclusion.tex
\section{Conclusion}
In this paper, we address two major limitations of existing Novel Class Discovery (NCD) methods: the neglect of multi-view data and the reliance on pseudo-labels. To tackle these issues, we propose an Intra-view and Inter-view Correlation Guided Multi-view Novel Class Discovery method. By leveraging matrix factorization, we effectively capture the distributional consistency between different views and directly fuse the resulting factor matrices to generate class labels. Unlike traditional multi-view clustering methods that lack prior information, our approach utilizes supervision from known classes to guide the learning of view weights for unknown classes, significantly improving performance. Experimental results demonstrate the effectiveness of the proposed algorithm. To the best of our knowledge, this is the first attempt to apply NCD in the multi-view domain, providing valuable insights and inspiration for the future application of NCD methods in multi-view scenarios. In the future, we expect to research how to utilize known class label information to select hyperparameters for novel classes.

%% file: sec/7_acknowledgments.tex
\section{Acknowledgments}
 This work is supported in part by the National Science and Technology Innovation 2030 Major Project of China under Grant No. 2022ZD0209103; in part by Major Program Project of Xiangjiang Laboratory under Grant No. 24XJJCYJ01002; in part by National Natural Science Foundation of China under Grant No. 62476281, 62306324, 62376279; in part by National Natural Science Foundation of China Joint Found under Grant No. U24A20323, U24A20333; in part by the Science and Technology Innovation Program of Hunan Province under Grant No. 2024RC3128; and in part by National University of Defense Technology Research Foundation under Grant No. ZK24-30.

%% file: main.bbl
\begin{thebibliography}{67}
\providecommand{\natexlab}[1]{#1}
\providecommand{\url}[1]{\texttt{#1}}
\expandafter\ifx\csname urlstyle\endcsname\relax
  \providecommand{\doi}[1]{doi: #1}\else
  \providecommand{\doi}{doi: \begingroup \urlstyle{rm}\Url}\fi

\bibitem[Chang et~al.(2017)Chang, Wang, Meng, Xiang, and Pan]{chang2017deep}
Jianlong Chang, Lingfeng Wang, Gaofeng Meng, Shiming Xiang, and Chunhong Pan.
\newblock Deep adaptive image clustering.
\newblock In \emph{Proceedings of the IEEE international conference on computer vision}, pages 5879--5887, 2017.

\bibitem[Chen et~al.(2020)Chen, Huang, Wang, and Huang]{2020Multi}
M.~S. Chen, L. Huang, C.~D. Wang, and D. Huang.
\newblock Multi-view clustering in latent embedding space.
\newblock In \emph{The 34th AAAI Conference on Artificial Intelligence}, 2020.

\bibitem[Dang et~al.(2021)Dang, Deng, Yang, Wei, and Huang]{dang2021nearest}
Zhiyuan Dang, Cheng Deng, Xu Yang, Kun Wei, and Heng Huang.
\newblock Nearest neighbor matching for deep clustering.
\newblock In \emph{Proceedings of the IEEE/CVF conference on computer vision and pattern recognition}, pages 13693--13702, 2021.

\bibitem[Dong et~al.(2023)Dong, Wang, Jin, Liu, and Zhu]{dong2023cross}
Zhibin Dong, Siwei Wang, Jiaqi Jin, Xinwang Liu, and En Zhu.
\newblock Cross-view topology based consistent and complementary information for deep multi-view clustering.
\newblock In \emph{Proceedings of the IEEE/CVF International Conference on Computer Vision}, pages 19440--19451, 2023.

\bibitem[Dong et~al.(2025)Dong, Liu, Wang, Liang, Zhang, Liu, Jin, Liu, and Zhu]{dong2025enhanced}
Zhibin Dong, Meng Liu, Siwei Wang, Ke Liang, Yi Zhang, Suyuan Liu, Jiaqi Jin, Xinwang Liu, and En Zhu.
\newblock Enhanced then progressive fusion with view graph for multi-view clustering.
\newblock In \emph{Proceedings of the Computer Vision and Pattern Recognition Conference}, pages 15518--15527, 2025.

\bibitem[Fini et~al.(2021)Fini, Sangineto, Lathuili{\`e}re, Zhong, Nabi, and Ricci]{fini2021unified}
Enrico Fini, Enver Sangineto, St{\'e}phane Lathuili{\`e}re, Zhun Zhong, Moin Nabi, and Elisa Ricci.
\newblock A unified objective for novel class discovery.
\newblock In \emph{Proceedings of the IEEE/CVF international conference on computer vision}, pages 9284--9292, 2021.

\bibitem[Gao et~al.(2019)Gao, Yu, Jin, and Yin]{Gao2019MultiviewLM}
Shengxiang Gao, Zhengtao Yu, Taisong Jin, and Ming Yin.
\newblock Multi-view low-rank matrix factorization using multiple manifold regularization.
\newblock \emph{Neurocomputing}, 335:\penalty0 143--152, 2019.

\bibitem[Gao et~al.(2024)Gao, Yi, Qin, Ye, Zhu, and Xu]{gao2024learning}
Zhirui Gao, Renjiao Yi, Zheng Qin, Yunfan Ye, Chenyang Zhu, and Kai Xu.
\newblock Learning accurate template matching with differentiable coarse-to-fine correspondence refinement.
\newblock \emph{Computational Visual Media}, 10\penalty0 (2):\penalty0 309--330, 2024.

\bibitem[Gao et~al.(2025)Gao, Yi, Zhu, Zhuang, Chen, and Xu]{gao2025generic}
Zhirui Gao, Renjiao Yi, Chenyang Zhu, Ke Zhuang, Wei Chen, and Kai Xu.
\newblock Generic objects as pose probes for few-shot view synthesis.
\newblock \emph{IEEE Transactions on Circuits and Systems for Video Technology}, 2025.

\bibitem[Gu et~al.(2023)Gu, Zhang, Xu, and He]{10376556}
Peiyan Gu, Chuyu Zhang, Ruijie Xu, and Xuming He.
\newblock Class-relation knowledge distillation for novel class discovery.
\newblock In \emph{2023 IEEE/CVF International Conference on Computer Vision (ICCV)}, pages 16428--16437, 2023.

\bibitem[Han et~al.(2019)Han, Vedaldi, and Zisserman]{Han_2019_ICCV}
Kai Han, Andrea Vedaldi, and Andrew Zisserman.
\newblock Learning to discover novel visual categories via deep transfer clustering.
\newblock In \emph{Proceedings of the IEEE/CVF International Conference on Computer Vision (ICCV)}, 2019.

\bibitem[Han et~al.(2020)Han, Rebuffi, Ehrhardt, Vedaldi, and Zisserman]{hanautomatically}
Kai Han, Sylvestre-Alvise Rebuffi, Sebastien Ehrhardt, Andrea Vedaldi, and Andrew Zisserman.
\newblock Automatically discovering and learning new visual categories with ranking statistics.
\newblock In \emph{International Conference on Learning Representations}, 2020.

\bibitem[Hou et~al.(2024)Hou, Chen, Zhu, Liu, Shi, Liu, Wu, and Xu]{hou2024nc2d}
Yue Hou, Xueyuan Chen, He Zhu, Ruomei Liu, Bowen Shi, Jiaheng Liu, Junran Wu, and Ke Xu.
\newblock Nc2d: Novel class discovery for node classification.
\newblock In \emph{Proceedings of the 33rd ACM International Conference on Information and Knowledge Management}, pages 849--859, 2024.

\bibitem[Hruschka et~al.(2009)Hruschka, Campello, Freitas, et~al.]{hruschka2009survey}
Eduardo~Raul Hruschka, Ricardo~JGB Campello, Alex~A Freitas, et~al.
\newblock A survey of evolutionary algorithms for clustering.
\newblock \emph{IEEE Transactions on systems, man, and cybernetics, Part C (applications and reviews)}, 39\penalty0 (2):\penalty0 133--155, 2009.

\bibitem[Hsu et~al.(2018)Hsu, Lv, and Kira]{hsu2018learning}
Yen-Chang Hsu, Zhaoyang Lv, and Zsolt Kira.
\newblock Learning to cluster in order to transfer across domains and tasks.
\newblock In \emph{International Conference on Learning Representations}, 2018.

\bibitem[Hsu et~al.(2019)Hsu, Lv, Schlosser, Odom, and Kira]{hsumulti}
Yen-Chang Hsu, Zhaoyang Lv, Joel Schlosser, Phillip Odom, and Zsolt Kira.
\newblock Multi-class classification without multi-class labels.
\newblock In \emph{International Conference on Learning Representations}, 2019.

\bibitem[Jain(2010)]{jain2010data}
Anil~K Jain.
\newblock Data clustering: 50 years beyond k-means.
\newblock \emph{Pattern recognition letters}, 31\penalty0 (8):\penalty0 651--666, 2010.

\bibitem[Jin et~al.(2023)Jin, Wang, Dong, Liu, and Zhu]{jin2023deep}
Jiaqi Jin, Siwei Wang, Zhibin Dong, Xinwang Liu, and En Zhu.
\newblock Deep incomplete multi-view clustering with cross-view partial sample and prototype alignment.
\newblock In \emph{Proceedings of the IEEE/CVF conference on computer vision and pattern recognition}, pages 11600--11609, 2023.

\bibitem[Joseph et~al.(2022)Joseph, Paul, Aggarwal, Biswas, Rai, Han, and Balasubramanian]{joseph2022novel}
KJ Joseph, Sujoy Paul, Gaurav Aggarwal, Soma Biswas, Piyush Rai, Kai Han, and Vineeth~N Balasubramanian.
\newblock Novel class discovery without forgetting.
\newblock In \emph{European Conference on Computer Vision}, pages 570--586. Springer, 2022.

\bibitem[Kang et~al.(2019)Kang, Zhou, Zhao, Shao, Han, and Xu]{LMVSC}
Zhao Kang, Wangtao Zhou, Zhitong Zhao, Junming Shao, Meng Han, and Zenglin Xu.
\newblock Large-scale multi-view subspace clustering in linear time.
\newblock In \emph{AAAI Conference on Artificial Intelligence}, 2019.

\bibitem[Li et~al.(2023{\natexlab{a}})Li, Zhang, Wang, Liu, Li, and Li]{LiLiangTKDE23}
Liang Li, Junpu Zhang, Siwei Wang, Xinwang Liu, Kenli Li, and Keqin Li.
\newblock Multi-view bipartite graph clustering with coupled noisy feature filter.
\newblock \emph{IEEE Transactions on Knowledge and Data Engineering}, 35\penalty0 (12):\penalty0 12842--12854, 2023{\natexlab{a}}.

\bibitem[Li et~al.(2024)Li, Pan, Liu, Liu, Liu, Li, Tsang, and Li]{LiLangTKDE24}
Liang Li, Yuangang Pan, Jie Liu, Yue Liu, Xinwang Liu, Kenli Li, Ivor~W. Tsang, and Keqin Li.
\newblock Bgae: Auto-encoding multi-view bipartite graph clustering.
\newblock \emph{IEEE Transactions on Knowledge and Data Engineering}, 36\penalty0 (8):\penalty0 1--14, 2024.

\bibitem[Li et~al.(2023{\natexlab{b}})Li, Liu, Zhang, and Liang]{10011211}
Miaomiao Li, Xinwang Liu, Yi Zhang, and Weixuan Liang.
\newblock Late fusion multiview clustering via min-max optimization.
\newblock \emph{IEEE Transactions on Neural Networks and Learning Systems}, pages 1--11, 2023{\natexlab{b}}.

\bibitem[Li et~al.(2023{\natexlab{c}})Li, Fan, Huo, and Gao]{10203164}
Wenbin Li, Zhichen Fan, Jing Huo, and Yang Gao.
\newblock Modeling inter-class and intra-class constraints in novel class discovery.
\newblock In \emph{2023 IEEE/CVF Conference on Computer Vision and Pattern Recognition (CVPR)}, pages 3449--3458, 2023{\natexlab{c}}.

\bibitem[Liu et~al.(2013)Liu, Wang, Gao, and Han]{liu2013multi}
Jialu Liu, Chi Wang, Jing Gao, and Jiawei Han.
\newblock Multi-view clustering via joint nonnegative matrix factorization.
\newblock In \emph{Proceedings of the 2013 SIAM international conference on data mining}, pages 252--260. SIAM, 2013.

\bibitem[Liu et~al.(2021{\natexlab{a}})Liu, Liu, Yang, Guo, Kloft, and He]{liu2021multiview}
Jiyuan Liu, Xinwang Liu, Yuexiang Yang, Xifeng Guo, Marius Kloft, and Liangzhong He.
\newblock Multiview subspace clustering via co-training robust data representation.
\newblock \emph{IEEE Transactions on Neural Networks and Learning Systems}, 33\penalty0 (10):\penalty0 5177--5189, 2021{\natexlab{a}}.

\bibitem[Liu et~al.(2021{\natexlab{b}})Liu, Liu, Yang, Liu, Wang, Liang, and Shi]{OPMC}
Jiyuan Liu, Xinwang Liu, Yuexiang Yang, Li Liu, Siqi Wang, Weixuan Liang, and Jiangyong Shi.
\newblock One-pass multi-view clustering for large-scale data.
\newblock In \emph{Proceedings of the IEEE/CVF International Conference on Computer Vision (ICCV)}, pages 12344--12353, 2021{\natexlab{b}}.

\bibitem[Liu et~al.(2023{\natexlab{a}})Liu, Liu, Yang, Liao, and Xia]{liu2023contrastive}
Jiyuan Liu, Xinwang Liu, Yuexiang Yang, Qing Liao, and Yuanqing Xia.
\newblock Contrastive multi-view kernel learning.
\newblock \emph{IEEE Transactions on Pattern Analysis and Machine Intelligence}, 45\penalty0 (8):\penalty0 9552--9566, 2023{\natexlab{a}}.

\bibitem[Liu et~al.(2025)Liu, Liu, Li, Wan, Tan, Zhang, Liang, Qu, Feng, Guan, and Liang]{Liu_2025_CVPR}
Jiyuan Liu, Xinwang Liu, Chuankun Li, Xinhang Wan, Hao Tan, Yi Zhang, Weixuan Liang, Qian Qu, Yu Feng, Renxiang Guan, and Ke Liang.
\newblock Large-scale multi-view tensor clustering with implicit linear kernels.
\newblock In \emph{Proceedings of the Computer Vision and Pattern Recognition Conference (CVPR)}, pages 20727--20736, 2025.

\bibitem[Liu et~al.(2024{\natexlab{a}})Liu, Liang, Dong, Wang, Yang, Zhou, Zhu, and Liu]{AEVC}
Suyuan Liu, Ke Liang, Zhibin Dong, Siwei Wang, Xihong Yang, Sihang Zhou, En Zhu, and Xinwang Liu.
\newblock Learn from view correlation: An anchor enhancement strategy for multi-view clustering.
\newblock In \emph{2024 IEEE/CVF Conference on Computer Vision and Pattern Recognition (CVPR)}, pages 26151--26161, 2024{\natexlab{a}}.

\bibitem[Liu et~al.(2024{\natexlab{b}})Liu, Liao, Wang, Liu, and Zhu]{RCAGL}
Suyuan Liu, Qing Liao, Siwei Wang, Xinwang Liu, and En Zhu.
\newblock Robust and consistent anchor graph learning for multi-view clustering.
\newblock \emph{IEEE Transactions on Knowledge and Data Engineering}, 36\penalty0 (8):\penalty0 4207--4219, 2024{\natexlab{b}}.

\bibitem[Liu et~al.(2024{\natexlab{c}})Liu, Wang, Liang, Zhang, Dong, Liu, Zhu, Liu, and He]{liu2024alleviate}
Suyuan Liu, Siwei Wang, Ke Liang, Junpu Zhang, Zhibin Dong, Tianrui Liu, En Zhu, Xinwang Liu, and Kunlun He.
\newblock Alleviate anchor-shift: Explore blind spots with cross-view reconstruction for incomplete multi-view clustering.
\newblock \emph{Advances in Neural Information Processing Systems}, 37:\penalty0 87509--87531, 2024{\natexlab{c}}.

\bibitem[Liu et~al.(2017)Liu, Zhou, Wang, Li, Dou, Zhu, and Yin]{10.5555/3298483.3298566}
Xinwang Liu, Sihang Zhou, Yueqing Wang, Miaomiao Li, Yong Dou, En Zhu, and Jianping Yin.
\newblock Optimal neighborhood kernel clustering with multiple kernels.
\newblock In \emph{Proceedings of the Thirty-First AAAI Conference on Artificial Intelligence}, page 2266–2272. AAAI Press, 2017.

\bibitem[Liu et~al.(2022)Liu, Tu, Zhou, Liu, Song, Yang, and Zhu]{liu2022deep}
Yue Liu, Wenxuan Tu, Sihang Zhou, Xinwang Liu, Linxuan Song, Xihong Yang, and En Zhu.
\newblock Deep graph clustering via dual correlation reduction.
\newblock In \emph{Proceedings of the AAAI conference on artificial intelligence}, pages 7603--7611, 2022.

\bibitem[Liu et~al.(2023{\natexlab{b}})Liu, Yang, Zhou, Liu, Wang, Liang, Tu, and Li]{liu2023simple}
Yue Liu, Xihong Yang, Sihang Zhou, Xinwang Liu, Siwei Wang, Ke Liang, Wenxuan Tu, and Liang Li.
\newblock Simple contrastive graph clustering.
\newblock \emph{IEEE Transactions on Neural Networks and Learning Systems}, 2023{\natexlab{b}}.

\bibitem[Lu et~al.(2016)Lu, Yan, and Lin]{7451227}
Canyi Lu, Shuicheng Yan, and Zhouchen Lin.
\newblock Convex sparse spectral clustering: Single-view to multi-view.
\newblock \emph{IEEE Transactions on Image Processing}, 25\penalty0 (6):\penalty0 2833--2843, 2016.

\bibitem[Murtagh and Contreras(2012)]{murtagh2012algorithms}
Fionn Murtagh and Pedro Contreras.
\newblock Algorithms for hierarchical clustering: an overview.
\newblock \emph{Wiley Interdisciplinary Reviews: Data Mining and Knowledge Discovery}, 2\penalty0 (1):\penalty0 86--97, 2012.

\bibitem[Nie et~al.(2014)Nie, Wang, and Huang]{nie2014clustering}
Feiping Nie, Xiaoqian Wang, and Heng Huang.
\newblock Clustering and projected clustering with adaptive neighbors.
\newblock In \emph{Proceedings of the 20th ACM SIGKDD international conference on Knowledge discovery and data mining}, pages 977--986, 2014.

\bibitem[Peng et~al.(2019)Peng, Huang, Lv, Zhu, and Zhou]{pmlr-v97-peng19a}
Xi Peng, Zhenyu Huang, Jiancheng Lv, Hongyuan Zhu, and Joey~Tianyi Zhou.
\newblock {COMIC}: Multi-view clustering without parameter selection.
\newblock In \emph{Proceedings of the 36th International Conference on Machine Learning}, pages 5092--5101. PMLR, 2019.

\bibitem[Ren et~al.(2024)Ren, Pu, Yang, Xu, Li, Pu, Philip, and He]{ren2024deep}
Yazhou Ren, Jingyu Pu, Zhimeng Yang, Jie Xu, Guofeng Li, Xiaorong Pu, S~Yu Philip, and Lifang He.
\newblock Deep clustering: A comprehensive survey.
\newblock \emph{IEEE transactions on neural networks and learning systems}, 2024.

\bibitem[Roy et~al.(2022)Roy, Liu, Zhong, Sebe, and Ricci]{roy2022class}
Subhankar Roy, Mingxuan Liu, Zhun Zhong, Nicu Sebe, and Elisa Ricci.
\newblock Class-incremental novel class discovery.
\newblock In \emph{European Conference on Computer Vision}, pages 317--333. Springer, 2022.

\bibitem[Tu et~al.(2025)Tu, Zhou, Liu, Cai, Zhao, Liu, and He]{tu2025wage}
Wenxuan Tu, Sihang Zhou, Xinwang Liu, Zhiping Cai, Yawei Zhao, Yue Liu, and Kunlun He.
\newblock Wage: Weight-sharing attribute-missing graph autoencoder.
\newblock \emph{IEEE Transactions on Pattern Analysis and Machine Intelligence}, 2025.

\bibitem[Wan et~al.(2023)Wan, Liu, Liu, Wang, Wen, Liang, Zhu, Liu, and Zhou]{AWMVC}
Xinhang Wan, Xinwang Liu, Jiyuan Liu, Siwei Wang, Yi Wen, Weixuan Liang, En Zhu, Zhe Liu, and Lu Zhou.
\newblock Auto-weighted multi-view clustering for large-scale data.
\newblock In \emph{Proceedings of the Thirty-Seventh AAAI Conference on Artificial Intelligence and Thirty-Fifth Conference on Innovative Applications of Artificial Intelligence and Thirteenth Symposium on Educational Advances in Artificial Intelligence}. AAAI Press, 2023.

\bibitem[Wan et~al.(2024{\natexlab{a}})Wan, Liu, Gan, Liu, Wang, Wen, Wan, and Zhu]{10486880}
Xinhang Wan, Jiyuan Liu, Xinbiao Gan, Xinwang Liu, Siwei Wang, Yi Wen, Tianjiao Wan, and En Zhu.
\newblock One-step multi-view clustering with diverse representation.
\newblock \emph{IEEE Transactions on Neural Networks and Learning Systems}, pages 1--13, 2024{\natexlab{a}}.

\bibitem[Wan et~al.(2024{\natexlab{b}})Wan, Liu, Liu, Wen, Yu, Wang, Yu, Wan, Wang, and Zhu]{wan2024decouple}
Xinhang Wan, Jiyuan Liu, Xinwang Liu, Yi Wen, Hao Yu, Siwei Wang, Shengju Yu, Tianjiao Wan, Jun Wang, and En Zhu.
\newblock Decouple then classify: A dynamic multi-view labeling strategy with shared and specific information.
\newblock In \emph{Forty-first International Conference on Machine Learning}, 2024{\natexlab{b}}.

\bibitem[Wan et~al.(2024{\natexlab{c}})Wan, Liu, Yu, Qu, Li, Liu, Liang, Dong, and Zhu]{wan2024contrastive}
Xinhang Wan, Jiyuan Liu, Hao Yu, Qian Qu, Ao Li, Xinwang Liu, Ke Liang, Zhibin Dong, and En Zhu.
\newblock Contrastive continual multiview clustering with filtered structural fusion.
\newblock \emph{IEEE Transactions on Neural Networks and Learning Systems}, 2024{\natexlab{c}}.

\bibitem[Wan et~al.(2024{\natexlab{d}})Wan, Xiao, Liu, Liu, Liang, and Zhu]{10506102}
Xinhang Wan, Bin Xiao, Xinwang Liu, Jiyuan Liu, Weixuan Liang, and En Zhu.
\newblock Fast continual multi-view clustering with incomplete views.
\newblock \emph{IEEE Transactions on Image Processing}, 33:\penalty0 2995--3008, 2024{\natexlab{d}}.

\bibitem[Wang et~al.(2018)Wang, Tian, Yu, Liu, Zhan, and Wang]{8030316}
Jing Wang, Feng Tian, Hongchuan Yu, Chang~Hong Liu, Kun Zhan, and Xiao Wang.
\newblock Diverse non-negative matrix factorization for multiview data representation.
\newblock \emph{IEEE Transactions on Cybernetics}, 48\penalty0 (9):\penalty0 2620--2632, 2018.

\bibitem[Wang et~al.(2023)Wang, Tang, Wan, Zhang, Sun, and Zomaya]{wang2023efficient}
Jun Wang, Chang Tang, Zhiguo Wan, Wei Zhang, Kun Sun, and Albert~Y Zomaya.
\newblock Efficient and effective one-step multiview clustering.
\newblock \emph{IEEE Transactions on Neural Networks and Learning Systems}, 2023.

\bibitem[Wang et~al.(2024)Wang, Li, Tang, Liu, Wan, and Liu]{wang2024multiple}
Jun Wang, Zhenglai Li, Chang Tang, Suyuan Liu, Xinhang Wan, and Xinwang Liu.
\newblock Multiple kernel clustering with adaptive multi-scale partition selection.
\newblock \emph{IEEE Transactions on Knowledge and Data Engineering}, 2024.

\bibitem[Wang et~al.(2019)Wang, Liu, Zhu, Tang, Liu, Hu, Xia, and Yin]{ijcai2019-524}
Siwei Wang, Xinwang Liu, En Zhu, Chang Tang, Jiyuan Liu, Jingtao Hu, Jingyuan Xia, and Jianping Yin.
\newblock Multi-view clustering via late fusion alignment maximization.
\newblock In \emph{Proceedings of the Twenty-Eighth International Joint Conference on Artificial Intelligence, {IJCAI-19}}, pages 3778--3784. International Joint Conferences on Artificial Intelligence Organization, 2019.

\bibitem[Wang et~al.(2022)Wang, Liu, Zhu, Zhang, Zhang, Gao, and Zhu]{FPMVS-CAG}
Siwei Wang, Xinwang Liu, Xinzhong Zhu, Pei Zhang, Yi Zhang, Feng Gao, and En Zhu.
\newblock Fast parameter-free multi-view subspace clustering with consensus anchor guidance.
\newblock \emph{IEEE Transactions on Image Processing}, 31:\penalty0 556--568, 2022.

\bibitem[Wen et~al.(2023)Wen, Wang, Liao, Liang, Liang, Wan, and Liu]{wen2023unpaired}
Yi Wen, Siwei Wang, Qing Liao, Weixuan Liang, Ke Liang, Xinhang Wan, and Xinwang Liu.
\newblock Unpaired multi-view graph clustering with cross-view structure matching.
\newblock \emph{IEEE Transactions on Neural Networks and Learning Systems}, 2023.

\bibitem[Wen et~al.(2025)Wen, Liu, Xu, Luo, Jia, Wu, Wang, Liang, Wang, Wang, et~al.]{wen2025measure}
Yi Wen, Yue Liu, Derong Xu, Huishi Luo, Pengyue Jia, Yiqing Wu, Siwei Wang, Ke Liang, Maolin Wang, Yiqi Wang, et~al.
\newblock Measure domain's gap: A similar domain selection principle for multi-domain recommendation.
\newblock \emph{arXiv preprint arXiv:2505.20227}, 2025.

\bibitem[Yang et~al.(2021)Yang, Zhang, Nie, Wang, Yu, and Wang]{9305974}
Ben Yang, Xuetao Zhang, Feiping Nie, Fei Wang, Weizhong Yu, and Rong Wang.
\newblock Fast multi-view clustering via nonnegative and orthogonal factorization.
\newblock \emph{IEEE Transactions on Image Processing}, 30:\penalty0 2575--2586, 2021.

\bibitem[Yao et~al.(2021)Yao, Li, Jiang, and Chen]{9212617}
Yaqiang Yao, Yang Li, Bingbing Jiang, and Huanhuan Chen.
\newblock Multiple kernel k-means clustering by selecting representative kernels.
\newblock \emph{IEEE Transactions on Neural Networks and Learning Systems}, 32\penalty0 (11):\penalty0 4983--4996, 2021.

\bibitem[Yu et~al.(2024{\natexlab{a}})Yu, Liang, Hu, Tu, Ma, Zhou, and Liu]{yu2024zoo}
Hao Yu, Ke Liang, Dayu Hu, Wenxuan Tu, Chuan Ma, Sihang Zhou, and Xinwang Liu.
\newblock Gzoo: Black-box node injection attack on graph neural networks via zeroth-order optimization.
\newblock \emph{IEEE Transactions on Knowledge and Data Engineering}, 2024{\natexlab{a}}.

\bibitem[Yu et~al.(2025{\natexlab{a}})Yu, Liang, LIANG, Liu, Liu, and Liu]{yu2025on}
Hao Yu, Weixuan Liang, KE LIANG, Suyuan Liu, Meng Liu, and Xinwang Liu.
\newblock On the adversarial robustness of multi-kernel clustering.
\newblock In \emph{{ICML}}, 2025{\natexlab{a}}.

\bibitem[Yu et~al.(2024{\natexlab{b}})Yu, Dong, Wang, Wan, Liu, Liang, Zhang, Tu, and Liu]{yu2024towards}
Shengju Yu, Zhibing Dong, Siwei Wang, Xinhang Wan, Yue Liu, Weixuan Liang, Pei Zhang, Wenxuan Tu, and Xinwang Liu.
\newblock Towards resource-friendly, extensible and stable incomplete multi-view clustering.
\newblock In \emph{Proceedings of the 41st International Conference on Machine Learning}, pages 57415--57440, 2024{\natexlab{b}}.

\bibitem[Yu et~al.(2024{\natexlab{c}})Yu, Wang, Wen, Wang, Luo, Zhu, and Liu]{10243081}
Shengju Yu, Siwei Wang, Yi Wen, Ziming Wang, Zhigang Luo, En Zhu, and Xinwang Liu.
\newblock How to construct corresponding anchors for incomplete multiview clustering.
\newblock \emph{IEEE Transactions on Circuits and Systems for Video Technology}, 34\penalty0 (4):\penalty0 2845--2860, 2024{\natexlab{c}}.

\bibitem[Yu et~al.(2025{\natexlab{b}})Yu, Liu, Wang, Tang, Luo, Liu, and Zhu]{10325611}
Shengju Yu, Suyuan Liu, Siwei Wang, Chang Tang, Zhigang Luo, Xinwang Liu, and En Zhu.
\newblock Sparse low-rank multi-view subspace clustering with consensus anchors and unified bipartite graph.
\newblock \emph{IEEE Transactions on Neural Networks and Learning Systems}, 36\penalty0 (1):\penalty0 1438--1452, 2025{\natexlab{b}}.

\bibitem[Zhang et~al.(2025{\natexlab{a}})Zhang, Wang, Jia, Li, Chen, and Li]{zhang2025multi}
Chao Zhang, Zhi Wang, Xiuyi Jia, Zechao Li, Chunlin Chen, and Huaxiong Li.
\newblock Multi-view clustering with incremental instances and views.
\newblock \emph{IEEE Transactions on Image Processing}, 2025{\natexlab{a}}.

\bibitem[Zhang et~al.(2021)Zhang, Liu, Wang, Liu, Dai, and Zhu]{10.1145/3474085.3475204}
Yi Zhang, Xinwang Liu, Siwei Wang, Jiyuan Liu, Sisi Dai, and En Zhu.
\newblock One-stage incomplete multi-view clustering via late fusion.
\newblock In \emph{Proceedings of the 29th ACM International Conference on Multimedia}, page 2717–2725, New York, NY, USA, 2021. Association for Computing Machinery.

\bibitem[Zhang et~al.(2025{\natexlab{b}})Zhang, Lin, Yan, Yao, Wan, Li, Zhang, Ke, and Xu]{Zhang_Lin_Yan_Yao_Wan_Li_Zhang_Ke_Xu_2025}
Yuanyang Zhang, Yijie Lin, Weiqing Yan, Li Yao, Xinhang Wan, Guangyuan Li, Chao Zhang, Guanzhou Ke, and Jie Xu.
\newblock Incomplete multi-view clustering via diffusion contrastive generation.
\newblock \emph{Proceedings of the AAAI Conference on Artificial Intelligence}, 39\penalty0 (21):\penalty0 22650--22658, 2025{\natexlab{b}}.

\bibitem[Zhang et~al.(2019)Zhang, Liu, Shen, Shen, and Shao]{8387526}
Zheng Zhang, Li Liu, Fumin Shen, Heng~Tao Shen, and Ling Shao.
\newblock Binary multi-view clustering.
\newblock \emph{IEEE Transactions on Pattern Analysis and Machine Intelligence}, 41\penalty0 (7):\penalty0 1774--1782, 2019.

\bibitem[Zhao et~al.(2022)Zhao, Zhong, Sebe, and Lee]{zhao2022novel}
Yuyang Zhao, Zhun Zhong, Nicu Sebe, and Gim~Hee Lee.
\newblock Novel class discovery in semantic segmentation.
\newblock In \emph{Proceedings of the IEEE/CVF conference on computer vision and pattern recognition}, pages 4340--4349, 2022.

\bibitem[Zhong et~al.(2021)Zhong, Fini, Roy, Luo, Ricci, and Sebe]{zhong2021neighborhood}
Zhun Zhong, Enrico Fini, Subhankar Roy, Zhiming Luo, Elisa Ricci, and Nicu Sebe.
\newblock Neighborhood contrastive learning for novel class discovery.
\newblock In \emph{Proceedings of the IEEE/CVF conference on computer vision and pattern recognition}, pages 10867--10875, 2021.

\end{thebibliography}
